\documentclass{article}

     \usepackage[preprint]{neurips_2025}

\usepackage[utf8]{inputenc} 
\usepackage[T1]{fontenc}    
\usepackage{hyperref}       
\usepackage{url}            
\usepackage{booktabs}       
\usepackage{amsfonts}       
\usepackage{nicefrac}       
\usepackage{microtype}      
\usepackage{xcolor}         
\usepackage{amsmath}
\usepackage{amsthm}
\usepackage{bm}
\usepackage{graphicx}
\usepackage{multirow}
\usepackage{multicol}
\usepackage{enumerate}
\usepackage{paralist}
\usepackage{subcaption}
\usepackage{tablefootnote}

\usepackage{booktabs}
\usepackage{xcolor}
\usepackage{dsfont}
\usepackage{algorithm}
\usepackage{algorithmic}
\usepackage{wrapfig}
\usepackage{pifont}

\title{Diffusion models under low-noise regime}

\author{
  Elizabeth Pavlova\\
  Department of Computer Science\\
  The University of Texas at Austin\\
  \texttt{elizabeth.pavlova@utexas.edu} \\
   \And
   Xue-Xin Wei\\
   Department of Neuroscience\\
   The University of Texas at Austin\\
   \texttt{weixx@utexas.edu} \\
}

\begin{document}

\maketitle

\begin{abstract}
Recent work on diffusion models proposed that they operate in two regimes: \emph{memorization}, in which models reproduce their training data, and \emph{generalization}, in which they generate novel samples. While this has been tested in high-noise settings, the behavior of diffusion models as effective denoisers when the corruption level is small remains unclear. To address this gap, we systematically investigated the behavior of diffusion models under low-noise diffusion dynamics, with implications for model robustness and interpretability. Using (i) CelebA subsets of varying sample sizes and (ii) analytic Gaussian mixture benchmarks, we reveal that models trained on disjoint data diverge near the data manifold even when their high-noise outputs converge. We quantify how training set size, data geometry, and model objective choice shape denoising trajectories and affect score accuracy, providing insights into how these models actually learn representations of data distributions. This work starts to address gaps in our understanding of generative model reliability in practical applications where small perturbations are common.
\end{abstract}

\section{Introduction}\label{sec:intro}
\vspace{-2mm}

Diffusion models have emerged as an effective class of deep generative models, delivering state-of-the-art results in image generation \cite{ho2020,dhariwal2021,rombach2022,saharia2022}, and more recently on video generation \cite{ho2022, singer2022}. These models operate by training a denoiser that estimates the score (gradient of the log density) of data distributions, then using it to sample from the corresponding estimated density through an iterative reverse denoising procedure \cite{ho2020, song2021, sohldickstein2015}.
Recent work proposes that such models exhibit two distinct operating regimes dictated by training dataset size. In the memorization regime~\cite{carlini2023, somepalli2022}, models trained on small datasets lead to overfitting, essentially reproducing training examples. In the generalization regime, larger datasets enable the model to generate novel but in-distribution samples \cite{kadkhodaie2023,halder2025}. The results in \cite{kadkhodaie2023} demonstrate that two denoisers trained on sufficiently large non-overlapping sets converge to similar denoising functions, providing direct evidence of generalization.

However, prior results on the regimes of generalization and memorization focused on large Gaussian noise initializations where the input of the denoiser was far from the data manifold. While these results suggest that score fields learned by the two models trained on disjoint sets of data indeed have similar large-scale structure, the extent to which the two score fields learned from disjoint datasets are \textit{locally} consistent remains unclear. Thus, there remains a fundamental gap in our understanding of diffusion model behavior in low-noise settings, precisely the conditions that arise in many practical applications such as denoising medical images \cite{pfaff2024,khader2023, mayo2024,kazerouni2023}, adversarial purification \cite{nie2022,lee2023robust}, and fine control of the content of the images generated \cite{zhang2023control, dhariwal2021,rombach2022}. 


In this paper, 
we investigate four questions that are pertinent for understanding the local behavior of diffusion models:
(1) How well do diffusion models trained on disjoint datasets generalize locally? We find trajectory divergence at small noise levels, even when models produce visually similar outputs at higher noise levels.
(2) How effectively can they denoise small perturbations? We discover that models struggle with denoising very small perturbations, with performance degrading as noise approaches zero.
(3) Do diffusion models exhibit attractor-like dynamics? Our novel "re-denoising" procedure reveals that small-dataset (but not large-dataset) models encode training images as discrete attractors, consistently returning to the same samples when perturbed.
(4) How accurately do they learn ground-truth score functions? Experiments with synthetic Gaussian mixtures show that current models struggle with complex geometric structures, favoring direct paths to density centers over geometrically accurate trajectories. Through these new analyses, we devised a set of evaluation methods that reveal the deficiency of current diffusion models in learning the detailed structure of data distributions and point to directions for future improvements.



\vspace{-2mm}

\section{Background and relevant work}\label{sec:background}
\vspace{-2mm}

\subsection{Diffusion Models and Denoising}
\vspace{-2mm}

Diffusion probabilistic models generate data by learning to reverse a stochastic noising process. The forward process corrupts clean data $\mathbf{x}_{0}$ through additive Gaussian noise:
\[
\mathbf{y} = \mathbf{x}_{0} + \sigma\boldsymbol{\epsilon}, \quad \boldsymbol{\epsilon} \sim \mathcal{N}(0, \mathbf{I}), \quad \sigma \in [0, \sigma_{\max}].
\]
The reverse process then attempts to reconstruct $\mathbf{x}_{0}$ from noisy input $\mathbf{y}$,  through an iterative denoising process.
Score-based diffusion models use a neural network that is trained to estimate the score function $\nabla_{\mathbf{x}} \log p(\mathbf{x})$ of the data distribution, which is a vector field that points in the direction of increasing probability density. 

For standard score-based models, the score function is approximated directly. However, when data lies on low-dimensional manifolds, direct score estimation can be unstable in regions with low data density. To address this limitation, noise-conditioned score networks (NCSNs) \cite{song2020} were developed, where the model $s_{\theta}(\mathbf{x}, \sigma)$ is trained to estimate the score of the data distribution perturbed with different noise levels $\sigma$.

A common training objective is denoising score matching:
\[
\mathcal{L}(\theta) = \mathbb{E}_{\sigma}\mathbb{E}_{\mathbf{x}_0}\mathbb{E}_{\boldsymbol{\epsilon}}\left[\|s_{\theta}(\mathbf{x}_0 + \sigma\boldsymbol{\epsilon}, \sigma) - (-\frac{\boldsymbol{\epsilon}}{\sigma})\|_2^2\right]
\]

Once trained, the model performs iterative denoising updates:
\begin{align*}
\mathbf{d}_{t} &= -s_{\theta}(\mathbf{x}^{(t)}, \sigma_{t}) \quad \text{(denoising direction)},\\
\mathbf{x}^{(t+1)} &= \mathbf{x}^{(t)} + h_{t}\,\mathbf{d}_{t} + \gamma_{t}\,\mathbf{z}_{t}, \quad \mathbf{z}_{t} \sim \mathcal{N}(0, \mathbf{I}),
\end{align*}
where $h_{t} \propto \sigma_{t}^{2}$ and $\gamma_{t} \propto \sigma_{t}$ define the integration schedule.

\subsection{Relevant work}
\vspace{-2mm}

\paragraph{Memorization in image generation}
A growing body of work around generative models examines whether their high-quality synthesis comes at the cost of memorizing training data. Recent work shows that diffusion networks can reproduce verbatim training samples, as demonstrated by large-scale extraction and pixel-level replication attacks \cite{carlini2023,somepalli2022,wang2024}. Yet when the dataset is sufficiently large, two denoisers trained on disjoint subsets converge to an almost identical score field and generate novel images, signaling a memorization to generalization transition that can be quantified through a formal sample-complexity analysis \cite{kadkhodaie2023,halder2025}. Several newer studies further improve our understanding of diffusion model memorization, introducing diagnostic metrics, structural explanations, and mitigation strategies \cite{li2024,gu2025,ross2025,chen2024,ren2024}, yet little is known about their local behavior.

\vspace{-2mm}
\paragraph{Score-based generative models}
Modern diffusion models trace back to foundational work around reversing a non-equilibrium diffusion process \cite{sohldickstein2015}. Noise Conditional Score Networks introduced multi-scale score matching, DDPM provided a variational formulation, the continuous-time SDE angle joined training and sampling, and a dynamical-systems analysis further characterized the stability of trajectories \cite{song2020,ho2020,song2021,biroli2024}. Subsequent work has focused on both faster samplers, using deterministic implicit steps (DDIM) or high-order ODE solvers such as DPM-Solver and DEIS, and on theoretical guarantees that sampling is as easy as score learning, through finite-step convergence proofs \cite{song2022ddim,lu2022,zhang2023,chen2023,lee2023,chen2023score}.

\vspace{-2mm}

\paragraph{Denoising at small noise levels}
Accurate score estimation at low-noise levels enables inverse problems, Bayesian imaging, and adversarial purification. A pretrained denoiser can act as an implicit prior for de-blurring and super-resolution \cite{kadkhodaie2021}, a principle extended to full posterior sampling \cite{janati2024} and analyzed in the context of medical MRI reconstruction \cite{pfaff2024}. Adversarial defenses purify extremely low-noise perturbations by running only a few reverse steps \cite{nie2022}, while low-noise denoising still exposes memorization in 3-D medical synthesis \cite{dar2023}. These empirical findings connect to classical results on modern image filtering and regularization by denoising \cite{milanfar2013,romano2017}. Additionally, since coarse structure is mostly fixed in the early, high-noise stages of the generative diffusion trajectory, recent methods, such as ControlNet and T2i-Adapter, achieve fine-grained control by manipulating the final low-noise steps \cite{mou2024, zhang2023control}.

\vspace{-2mm}
\section{Methodology}\label{sec:method}
\vspace{-2mm}

In this paper, we systematically investigate the properties of three variations of diffusion models using both real and synthetic datasets based on a number of evaluation metrics. For clarity, we summarize the key methodological details in this section.

\vspace{-2mm}

\subsection{Datasets}
\label{sec:data}
\vspace{-2mm}


\paragraph{CelebA Subsets.} To assess generalization and attractor behavior across data scales, we construct two mutually exclusive subsets of the CelebA dataset following \citep{kadkhodaie2023}. Each subset is grayscale and downsampled to $80 \times 80$ resolution, with cardinalities of 1, 10, 100, 1000, 10,000, and 100,000 samples. 
We train denoising models on each mutually exclusive subset across all training sizes using different model objectives. All models share architecture and optimization hyperparameters, differing mainly in dataset content, size, and objective.
\vspace{-2mm}
\paragraph{Synthetic datasets} To evaluate score field accuracy against known analytic gradients, we constructed three synthetic 2D Gaussian Mixtures distributions: (i) \textsc{Uniform}: A mixture of well-separated isotropic Gaussians. (ii) \textsc{SharpCov}: A mixture with highly anisotropic covariances (long elliptical components). (iii) \textsc{Spiral}: A non-convex distribution with 12 elliptical components arranged in a spiral pattern.


Each distribution is sampled to form a training set of 100,000 points, which can then be further broken down into cardinalities of 1, 10, 100, 1000, 10,000, and 100,000 samples.

\vspace{-2mm}

\subsection{Model Objectives and Architectures}
\label{sec:diffusionmodel}
\vspace{-2mm}

We evaluate and compare three classes of diffusion models. Each of these models learns a denoising vector field via a different training objective.

\begin{description}
    \item[UNet Denoiser \cite{kadkhodaie2023}]. This model directly minimizes reconstruction loss in pixel space:
    \[
    \mathcal{L}_{\text{recon}} = \| \mathbf{x}_{0} - f_{\theta}(\mathbf{y}) \|_{2}^{2},
    \]
    where $f_{\theta}$ is a UNet conditioned on noise level range between 0 and 1.0. It outputs a clean estimate $\hat{\mathbf{x}}$ from the noisy input $\mathbf{y}$ and implicitly learns a denoising direction $\mathbf{d} = \hat{\mathbf{x}} - \mathbf{y}$ which is used in trajectory integration. Note that this method is not directly trained to match score functions. The matching of the score function is implicitly encouraged. 

    \item[NCSN (Noise-Conditional Score Network) \cite{song2020}]. This model minimizes the expected squared error between the predicted score and the true reverse-time gradient:
    \[
    \mathcal{L}_{\text{NCSN}} = \mathbb{E}_{\sigma}\,\mathbb{E}_{\mathbf{y}|\sigma} \left[ \left\| s_{\theta}(\mathbf{y}, \sigma) + \frac{1}{\sigma^{2}} (\mathbf{y} - \mathbf{x}_{0}) \right\|_{2}^{2} \right],
    \]
    training the network to predict $\nabla_{\mathbf{y}} \log p_{\sigma}(\mathbf{y})$ across a range of noise levels. The score function is evaluated during each reverse diffusion step.

    \item[SSM (Sliced Score Matching) \cite{song2020}]. This model estimates the score by minimizing a projection-based approximation to the Fisher divergence:
    \[
    \mathcal{L}_{\text{SSM}} = \mathbb{E}_{\mathbf{v} \sim \mathcal{N}(0, I)} \mathbb{E}_{\mathbf{y}} \left[ \mathbf{v}^{\top} \nabla_{\mathbf{y}} s_{\theta}(\mathbf{y}) + \frac{1}{2} \| s_{\theta}(\mathbf{y}) \|_{2}^{2} \right],
    \]
    where random directions $\mathbf{v}$ are sampled to slice through the score field. This objective does not require access to ground-truth clean data and focuses on capturing global score structure.
\end{description}

All models are trained on the same data distributions, using comparable network capacities and architectural choices.

\vspace{-2mm}

\subsection{Metrics for comparing two diffusion models}
\label{sec:metric}

\vspace{-2mm}

To evaluate how different models reconstruct trajectories from the same noisy initialization, we compare two models $A$ and $B$ with trajectories $\{\mathbf{x}^{(t)}_A\}_t$ and $\{\mathbf{x}^{(t)}_B\}_t$ across several metrics. We define:
\begin{itemize}
    \item $L_2$ divergence at step $t$: $D_t^{L2} = \lVert \mathbf{x}^{(t)}_A - \mathbf{x}^{(t)}_B \rVert_2$
    \item Cosine divergence of predicted scores: $D_t^{\cos} = 1 - \dfrac{\langle s_A^{(t)}, s_B^{(t)} \rangle}{\|s_A^{(t)}\|_2 \, \|s_B^{(t)}\|_2}$
    \item Distance to data manifold: $M_t = \min_{\mathbf{x} \in \mathcal{M}} \lVert \mathbf{x}^{(t)} - \mathbf{x} \rVert_2$
    \item Score approximation error (for synthetic data): $E_t^{\text{score}} = \lVert s_{\theta}(\mathbf{x}^{(t)}, \sigma_t) - \nabla_{\mathbf{x}} \log p_{\sigma_t}(\mathbf{x}^{(t)}) \rVert_2$
\end{itemize}

For Gaussian mixture models, we compute the analytical score function $\nabla_{\mathbf{x}} \log p_{\sigma}(\mathbf{x})$ at different noise levels $\sigma$ to provide a ground truth baseline. We analyze metric behavior across varying distances from distribution modes, binning points into three categories: close, mid-range, and far from centers.



\begin{figure}[b!]
    \centering
    \vspace{-2mm}

    \includegraphics[width=0.99\linewidth]{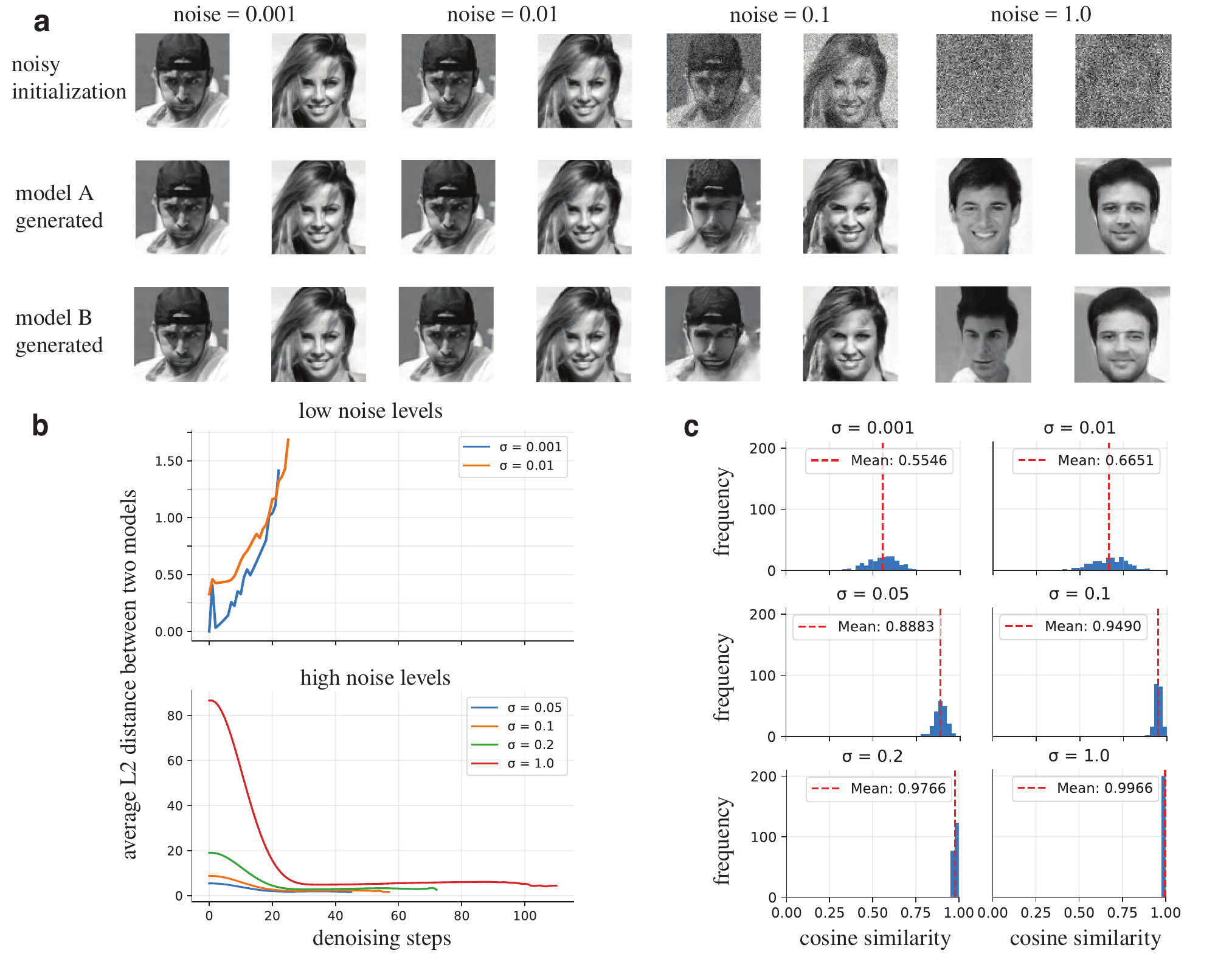}
    \caption{Probing the consistency of diffusion models trained on disjoint subsets of CelebA under varying noise levels. (a) Outputs of two models (A and B) trained on disjoint 100k CelebA subsets and denoising the same inputs across different initial noise levels. (b) Average $L_2$ distance between denoised trajectories of Models A and B, separated into low and high noise ranges. (c) Histogram of cosine similarity of denoising direction vectors for Models A and B at different noise levels.}
    \label{fig:Fig1}
\end{figure}

\vspace{-2mm}

\section{Results}\label{sec:experiments}
\vspace{-2mm}


We present a series of experiments addressing the four key questions outlined in the introduction, examining local generalization, performance on small perturbations, attractor dynamics, and accuracy in learning ground-truth score functions.

\vspace{-2mm}

\subsection{Diffusion models trained on disjoint datasets diverge at low-noise regime}\label{sec:exp:lownoise}
\vspace{-2mm}

Prior work has examined the ability of diffusion models to generalize. Kadkhodaie et al. \cite{kadkhodaie2023} split the CelebA dataset into two non-overlapping subsets and trained one model based on each subset. They found that with few training images the model memorizes, so iterative denoising from random noise collapses onto a training image. In contrast, when the training set is large, when initializing both models with the same random noise, the two models output similar images that are not from either training set. Note that these results were shown based on a particular UNet denoiser. We are interested in examining whether similar results hold for other popular diffusion models proposed. We followed the approach in \cite{kadkhodaie2023} to train two versions (i.e., Model A and Model B) of UNet denoiser developed in \cite{kadkhodaie2023} using on two non-overlapping subsets of CelebA, and we were able to replicate their main results. We then conducted some similar analyses on diffusion models we trained on NCSN and SSM objectives (see Section~\ref{sec:diffusionmodel}), and found similar results.
These results generalize the results in \cite{kadkhodaie2023} to a broader range of diffusion models.

The results above are based on initializing the network states at random noise, which can be interpreted as adding high noise to a clean image.
Geometrically, these results suggest that the score functions learned by diffusion models point toward the image manifold of faces, and that the global structures of these score functions are shared across two diffusion models trained on non-overlapping subsets of faces.



One important question is whether the two models are also consistent locally. To address this question, we developed a more general analysis framework. Instead of random noise initialization, we initialize the model at the vicinity of clean images. We then run denoising, and quantify whether the two models lead to a similar transformation of the initial image. We call this procedure a "\textit{local probe}" of diffusion models--by gradually decreasing the size of noise perturbation, one can examine the consistency of the local structure of two diffusion models at finer spatial scales. 

Using this \textit{local probe}, we find the two models trained on disjoint subsets of CelebA diverge in low-noise settings. While results in \cite{kadkhodaie2023} showed that models trained on large datasets (100k images) generate similar outputs from high noise initialization, we find that this convergence breaks down near the data manifold. Fig.\ref{fig:Fig1}a shows outputs from two UNet models trained on disjoint 100k CelebA subsets, denoising the same inputs across noise levels. Although outputs look visually similar across levels, quantification of the differences of the two below reveal divergence at low-noise (Fig.~\ref{fig:Fig1}b,c).

Specifically, we quantify the differences using two metrics run on 200 image samples initialized with different levels of Gaussian noise added. The first metric is the $L_2$ distance between denoised trajectories of both models at each timestep (see Fig.~\ref{fig:Fig1}b). The results suggest that for large noise, the distance between two models gradually decreases as denoising steps increases, consistent with results in \cite{kadkhodaie2023}. However, for small noise (which was not examined in previous work), the distance between two models diverges as denoising steps increase. The second metric is the cosine similarity of the predicted score vector (Fig.~\ref{fig:Fig1}c). When noise is large (e.g., $\sigma =1.0$), this is essentially identical to the analysis performed in \cite{kadkhodaie2023}, and indeed the two denoisers have high cosine similarities. However, with decreasing noise levels, the two denoisers become less consistent. At the smallest noise level we examined ($\sigma =0.001$), the two models only exhibit a modest correlation of 0.55.



We performed the same analysis on models trained with smaller train set (1, 10, 100, 1k, 10k images). These models exhibit substantially higher trajectory divergence than those trained on 100k images, across all noise levels (see Appendix).

\begin{figure}[t!]
    \centering
     \includegraphics[width=0.94\linewidth]{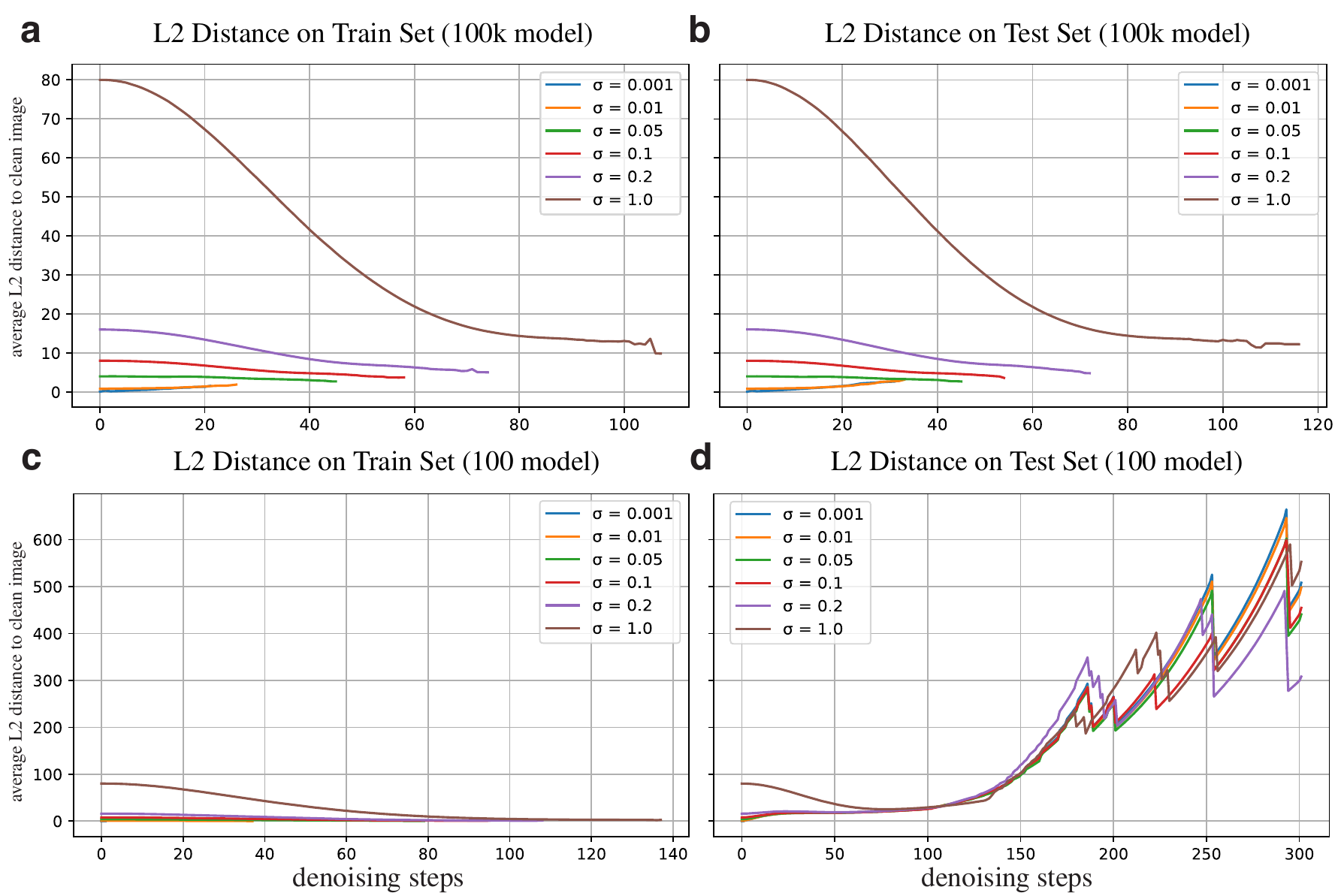}
   \caption{Denoising performance of diffusion models across noise levels. (a,b) Average $L_2$ distance from the clean image over denoising steps for the model trained on 100k images, shown on train and test sets, respectively. (c,d) Same analysis for the model trained on 100 images.}
   \vspace{-4mm}
    \label{fig:fig2}
\end{figure}

\vspace{-4mm}

\subsection{Denoising performance is limited at low noise levels}
\vspace{-2mm}

Diffusion models can function either as samplers when initialized from pure noise or as denoisers when initialized from a slightly corrupted image \cite{song2021, milanfar2013, romano2017}. 
Recently, diffusion models have been proposed to improve adversarial robustness by stripping away perturbations \cite{nie2022}. The idea is that, if a diffusion model can return the perturbed image back to (or closer to) the clean image, the classifier downstream will recover its clean-image accuracy \cite{lee2023robust}. It remains unclear how well diffusion models can remove such adversarial noise, which is often small. Given our earlier finding that the learned score function still shows local variability (even with 100k training images), we expect this variability to put a limit on how well diffusion models can remove local perturbations via denoising.

To test this, we trained a UNet denoiser on a large CelebA susbet (n=100k), and then evaluated the denoising performance for various levels of noise. Starting from the perturbed image, we track the $L_2$ distance to the original clean image. When the injected noise is substantial, this distance drops sharply (see Fig.~\ref{fig:fig2}a). However, at $\sigma=0.05$, this reduction becomes minimal, and at even lower noise levels the denoised output \textit{drifts further} \textit{away }from the clean target. The denoising performance for the training and test sets are comparable (Fig.~\ref{fig:fig2}a,b), consistent with the idea that diffusion models generalize when training set is large.

We also trained and analyzed a UNet denoiser on a small dataset (n= 100). For this model, the effect is split (Fig.~\ref{fig:fig2}c,d): it almost perfectly restores perturbed training images yet drifts even more on unseen test images, reflecting its pure memorization.

These results characterize a practical limit of diffusion models in removing local perturbations: local score variability and potential bias in the learned image prior. Practically, since $\sigma=0.05$ is  larger than the size of a typical adversarial attacks~\cite{Szegedy2013intriguing}, diffusion models may struggle to effectively push adversarial examples fully back to the clean manifold.

\vspace{-2mm}
\subsection{Models trained on small (but not large) datasets exhibit discrete-attractor dynamics}
\vspace{-2mm}

While there is evidence supporting the two regimes of diffusion models (memorization vs. generalization) depending on the size of the training set, it is mostly drawn from analyses conducted in the high-noise part of the reverse process.
Notably, classic neural network memory theory suggests that memories are stored as discrete attractors, e.g., a Hopfield network \cite{hopfield} represents each memory as a point attractor in high-dimensional space. One defining property of such discrete attractors is their resistance to local perturbations. Inspired by this theory, we examined whether diffusion models encode samples in the training set as discrete-attractors. Note that the results reported in Fig.~\ref{fig:fig2}c provide some preliminary support for discrete-attractor dynamics as the model trained on a small sample size exhibits systematic attraction towards the clean images. 

\begin{figure}[h!]
    \centering
    \includegraphics[width=0.99\linewidth]{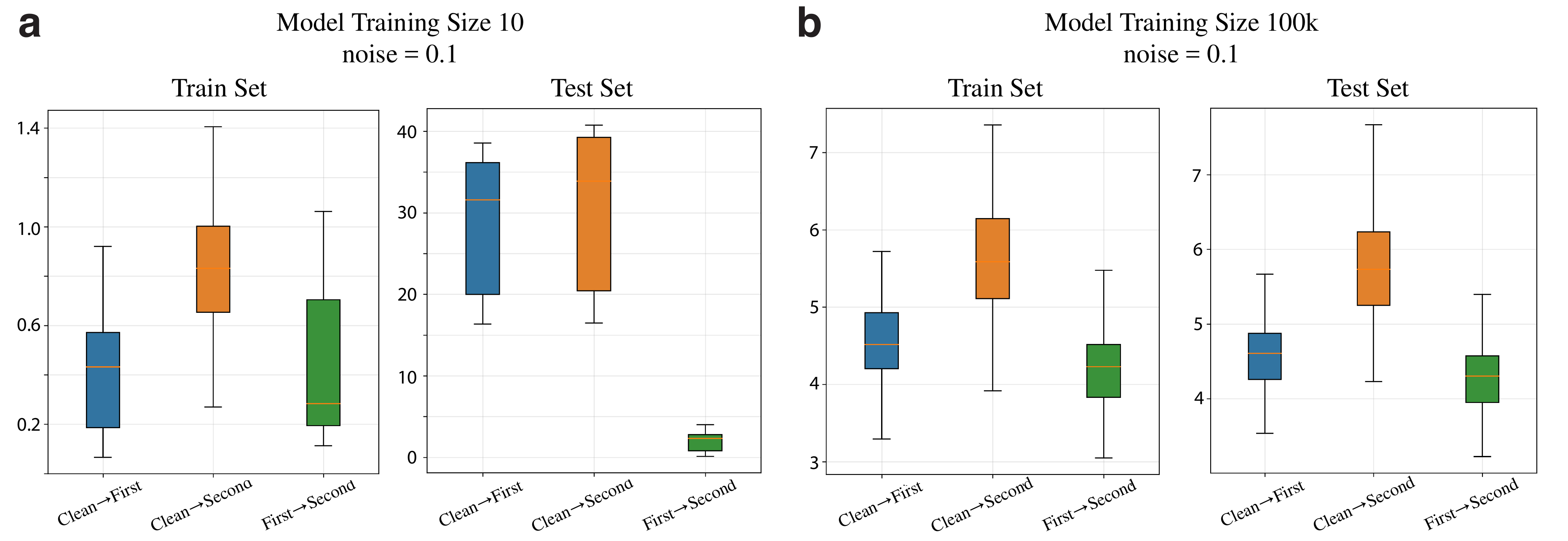}
    \caption{Stability of denoising trajectories under repeated passes. Boxplots show $L_2$ distances between (i) clean images and first denoised output, (ii) clean images and second denoised output, and (iii) first and second denoised outputs. (a): model trained on 10 images. (b): model trained on 100k images. 
    }

    \vspace{-3mm}
    \label{fig:fig3}
\end{figure}

To further investigate this, we train a UNet denoiser based on an even smaller sample size (n=10; memorization regime). We develop a novel "re-denoising" procedure to test the attractor dynamics. Specifically, for a given image, we add a small amount of noise and denoise it. We then add small noise again to the denoised output and perform denoising for the second time. Consistent convergence back to training images in this double-pass test signals the presence of a stable discrete-attractor. 

Fig. \ref{fig:fig3}a  shows the results of this analysis. When applying a local perturbation of $\sigma= 0.1$ (corresponding to an initial $L_2$ difference of 8.1 on average) to training set images, the images move towards clean images. After reapplying the perturbation, the images again move toward the original clean images. The average distance between the two denoised outputs is about 0.3. When repeating the same analysis on test set images, different patterns emerge. Denoising and re-denoising both lead to large distances to clean images. However, the distance between the two denoising outputs is small. This suggests that the perturbed image was initially denoised to an image in the training set, and then got attracted back to the same image during re-denoising.  
These results provide support for the hypothesis that diffusion models trained on small datasets encode training images as discrete-attractors. 


We apply analogous analysis on UNet denoiser trained on 100k images (Fig. \ref{fig:fig3}b). We find that, for both training and test sets, there is attraction toward clean images for a perturbation with $\sigma=0.1$. 
Notably, the distances between the first and second denoising outputs are substantially larger than those from the model trained on 10 images. These results suggest that the 100k-model does not encode the images as discrete attractors. Our interpretation is that the representation behaves more similarly to continuous attractors that smoothly interpolate between training images, allowing variability in a continuous space. The continuous-attractor-like representation may be important for generalization.



\vspace{-2mm}

\subsection{Diffusion models struggle to learn the local geometrical structure of 2D densities}\label{sec:exp:gmm}
\vspace{-2mm}

The analysis on natural image datasets can not directly inform us on the accuracy of the score functions learned by the diffusion models due to a lack of ground-truth. To address this limitation and directly quantify the accuracy of the score functions learned by diffusion models, we turn to simulated data. 
We design several synthetic datasets based on 2D Gaussian mixtures: (i) \textsc{Uniform} mixture; (ii) \textsc{SharpCov} mixture; (iii) \textsc{Spiral} mixture
(also see Section \ref{sec:data}). We evaluate diffusion models on these datasets with known score functions.

\begin{figure}[t]
    \centering
    \includegraphics[width=0.94\linewidth]{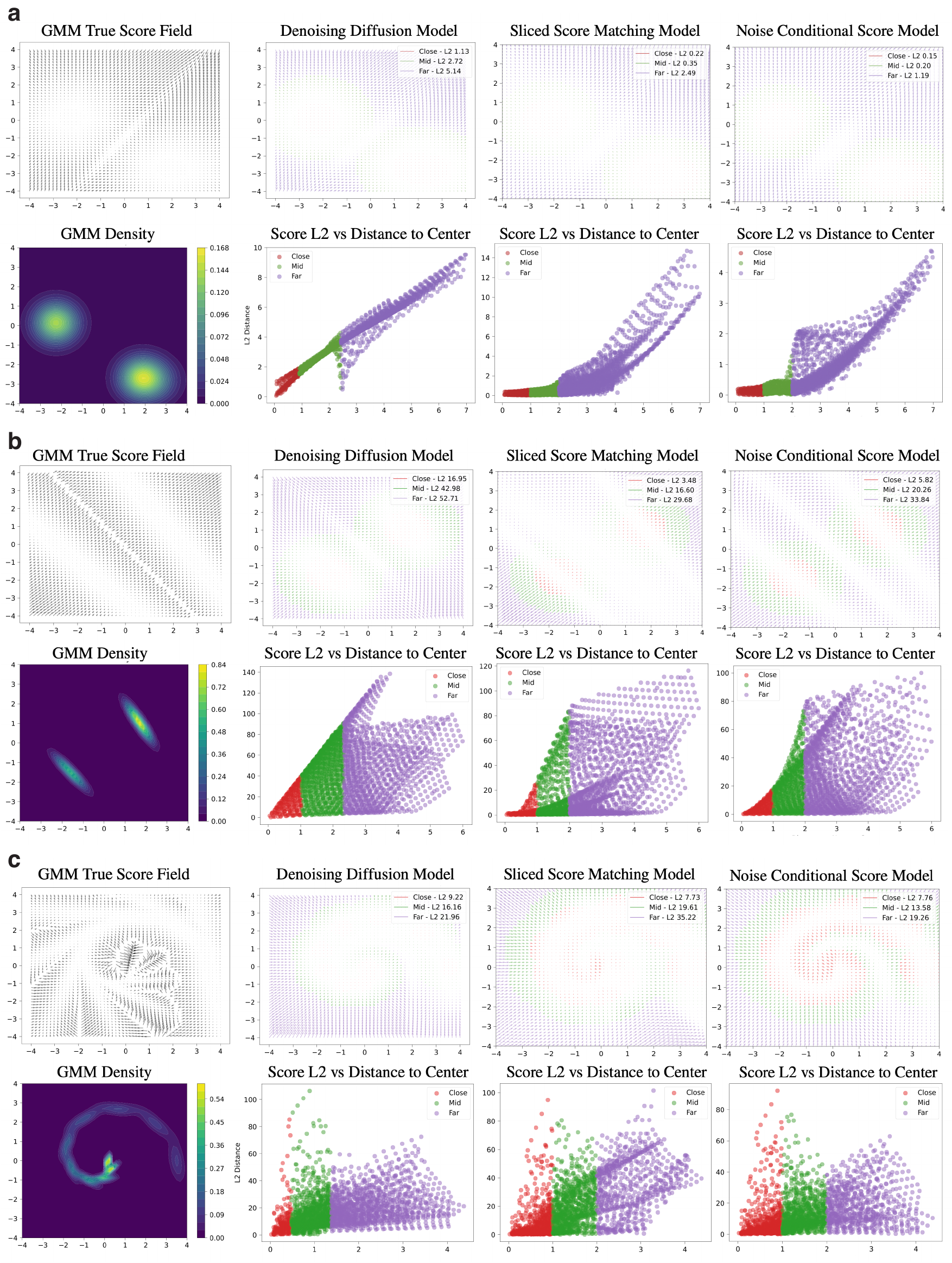}
    \caption{Score field accuracy comparison across synthetic Gaussian mixtures. Each pair of rows (a-c) shows a different mixture: (a) \textsc{Uniform}, (b) \textsc{SharpCov}, and (c) \textsc{Spiral}. For each mixture, the top row shows score fields (ground truth, then UNet, SSM, and NCSN models) and the bottom row shows GMM density (left) and $L_2$ error vs. distance plots for each model. Points in error plots and model score fields are color-coded by distance from distribution centers.}
    \label{fig:fig4}
    \vspace{-3mm}
\end{figure}

Fig.~\ref{fig:fig4} visualizes both the learned score fields and quantitative $L_2$ error compared to ground truth, across the three cases. These analyses revealed interesting failure modes of diffusion models. First, for the relatively simple case of \textsc{Uniform} mixture, we find that all three models recover the basic score structure well. Errors remain low even in sparse regions, indicating stable learning. Second, on the \textsc{SharpCov} mixture, UNet denoisers fail to track elongated score paths, especially in low-density tails. In contrast, NCSN and SSM significantly reduce score error in these regions, benefiting from explicit score-matching objectives. Third, on the \textsc{Spiral} mixture, all models exhibit a common failure mode. That is, their trajectories cut directly toward the mixture center, overshooting the curved manifold. This confirms a limitation in current architectures and objectives, which favor straight-line flows over geometrically accurate ones. 

\begin{figure}[h!]
    \centering
     \includegraphics[width=0.99\linewidth]{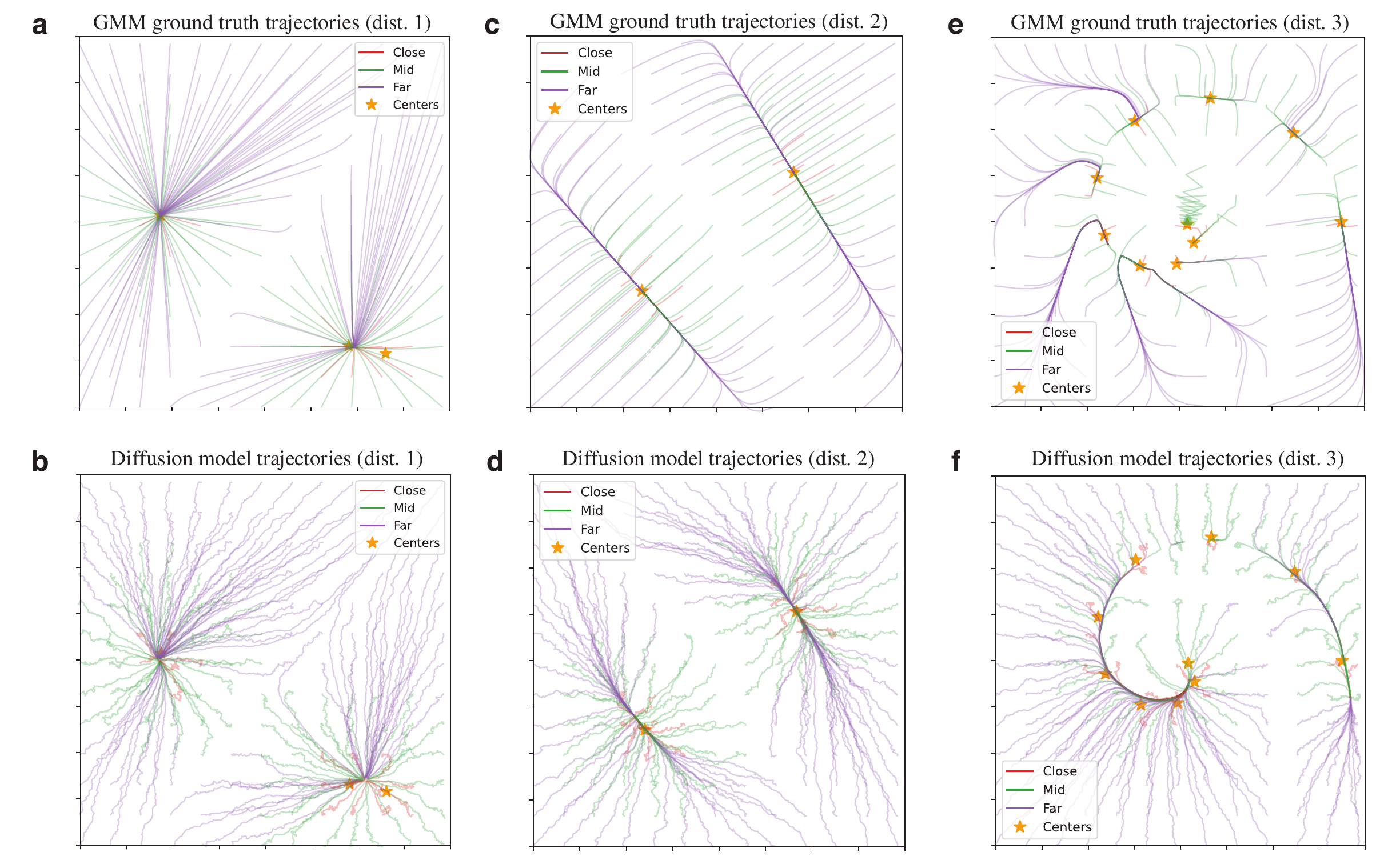}
   \caption{Full reverse diffusion trajectories on synthetic Gaussian mixtures. For each mixture, we visualize both the learned and true denoising trajectories to assess model alignment with the analytic gradient flow.}
    \label{fig:fig5}
\end{figure}

We then examine the denoising trajectories in their entirety to reveal how these errors accumulate during the full denoising process. Fig.~\ref{fig:fig5} compares ground truth trajectories (Fig.~\ref{fig:fig5}a,c,e) against model-generated trajectories (Fig.~\ref{fig:fig5}b,d,f) for each Gaussian mixture.
Across all three distributions, we observe that model trajectories successfully converge to high-density regions but often take different paths than the ground truth. In uniform mixtures, this difference is minimal. In distributions with sharp covariance, model trajectories fail to follow the principal axes as precisely as ground truth. Most notably, in spiral distributions, models create direct paths that don't preserve the curved geometry of the true manifold. The $L_2$ distance plots confirm this pattern, showing the same trend towards the centers across all three distributions.


These results are surprising, given the size of the synthetic datasets we used for learning these two-dimensional distribution is already quite large (n=100k). These results suggest that, while current diffusion models effectively locate density centers, they struggle to precisely capture the geometric structure of data manifolds, particularly when the fine spatial structure involves directional anisotropy or curvature.
Although the structure of natural image priors are not well understood, their local structure are surely more complicated than the 2-D datasets we have created. Given our results on simulated data, it is unlikely that current diffusion models can accurately learn the local structure of the CelebA dataset. This may help explain the trajectory divergence observed in our low-noise CelebA experiments, where subtle geometric differences between independently trained models become most apparent.

\section{Discussion}\label{sec:discussion}
\vspace{-3mm}

Our investigation into low-noise diffusion dynamics reveals several key insights about how these models represent data distributions. First, although two models trained on disjoint 100k image subsets generate similar outputs in the high-noise regime, we observe measurable divergence in their low-noise trajectories, implying they need not share identical local representations of the data manifold. Second, models trained on smaller datasets show substantially higher trajectory divergence across all noise levels, providing further evidence for the memorization hypothesis; the gap is largest at the lowest noise levels, closest to the data manifold. Third, In our synthetic tests, every architecture struggles with accurately representing complex geometric structures like anisotropic covariances and curved manifolds, highlighting a systematic biases in the score functions learned in diffusion models. Fourth, models trained with explicit score-matching objectives (NCSN, SSM) produce more accurate score fields in low-density regions compared to direct reconstruction objectives, suggesting the importance of objective choice for applications requiring accuracy far from the data manifold. These insights may inspire development of future generations of diffusion models that more faithfully learn the statistics of natural scene. 

{\bf Limitations}
Our work is based on the CelebA dataset and synthetic Gaussian mixture datasets, and it remains unclear to what extent the results would transfer to other image domains. Future work should extend these investigations to more diverse datasets, including natural scenes and medical images, to ensure relevance in applicable domains. Additionally, we focused on three specific model objectives, but the landscape of diffusion architectures continues to evolve. 
Further research could explore how architectural innovations affect low-noise behavior and whether they can address the challenges we identified.
Furthermore, we have only investigated diffusion models trained directly in pixel space. It would be interesting to investigate these questions for diffusion models trained in latent space.

\bibliographystyle{unsrt}
\bibliography{lowNoise}

\newpage
\section*{Appendix}

\section*{A. Broader impacts}
By analyzing when diffusion models succeed or fail at removing small perturbations, our study can help practitioners choose safer settings for medical imaging and robustness tasks; it may also inspire objectives that reduce training memorization and in turn privacy leakage. Conversely, the same insights could aid attackers in crafting perturbations that bypass diffusion-based defenses or in steering models to memorize sensitive content. 

\section*{B. Additional experimental results}

\subsection*{B1. Experimental setup}
To systematically compare trained models, we evaluate each across multiple noise levels $\sigma \in \{0.001, 0.01, 0.05, 0.1, 0.2, 1.0\}$. 

Denoising trajectories are always initialized from an identical corrupted input; the
only source of variation is the model’s learned vector field.  Reverse steps continue
until either (i) the change in output between consecutive iterations falls below a
$\sigma$-dependent threshold or (ii) a model-specific maximum step count is reached.

\subsection*{B2. Effect of training‐set size on trajectory consistency}

\begin{figure}[h!]
    \centering
     \includegraphics[width=0.83\linewidth]{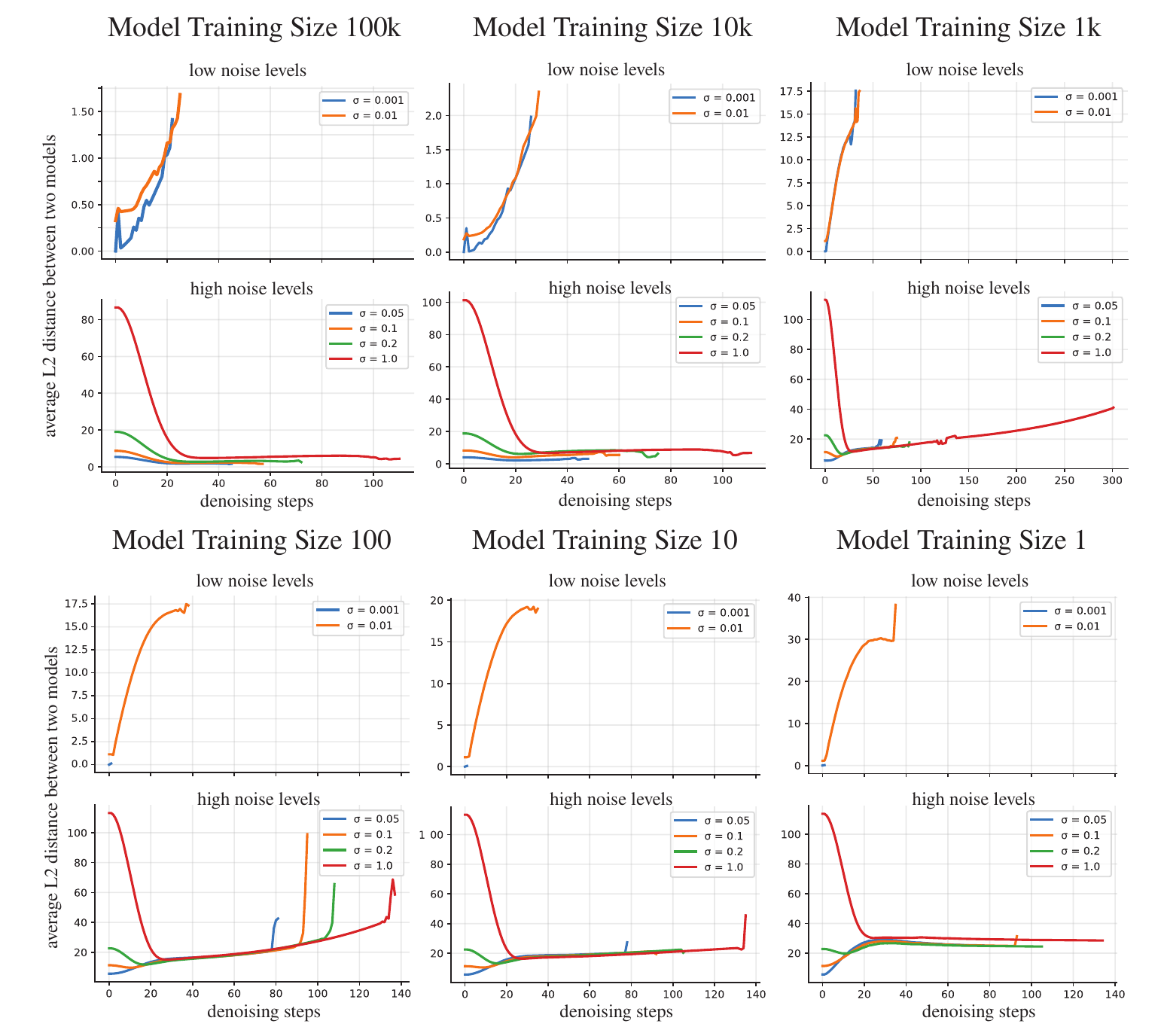}
   \caption{Effect of training set size on local consistency of diffusion models. Two models are trained on disjoint subsets of the same cardinality and evaluated on identical corrupted inputs. Average $L_2$ distance between denoised trajectories of Models A and B, separated into low and high noise ranges.}
    \label{fig:fig6}
\end{figure}

\begin{figure}[h!]
    \centering
     \includegraphics[width=0.99\linewidth]{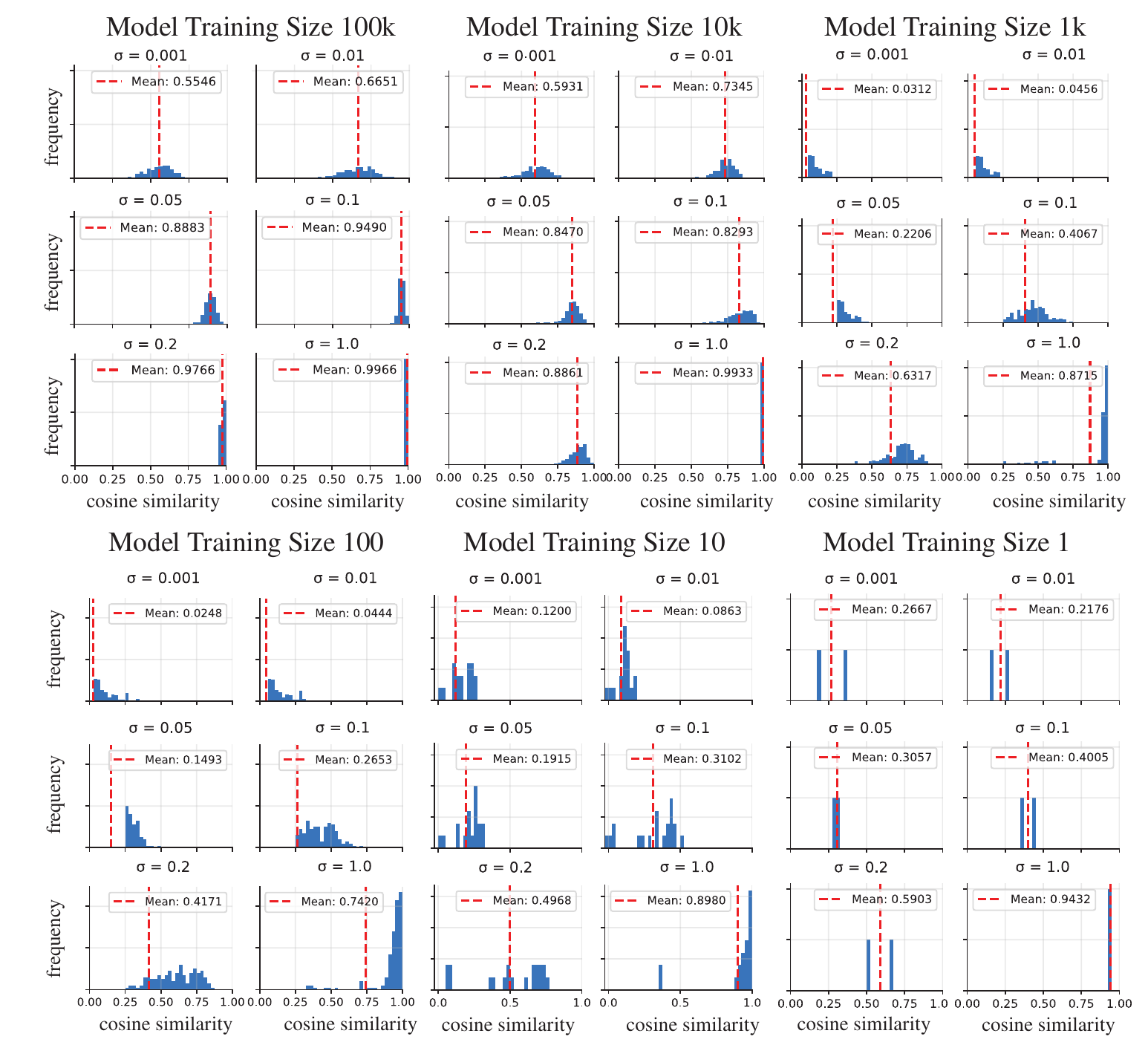}
   \caption{Effect of training set size on local consistency of diffusion models. Two models are trained on disjoint subsets of the same cardinality and evaluated on identical corrupted inputs. Histogram of cosine similarity of denoising direction vectors for Models A and B at different noise levels.}
    \label{fig:fig7}
\end{figure}

We next extend the divergence analysis of Figure \ref{fig:Fig1} to all CelebA subset sizes. Figures \ref{fig:fig6} and \ref{fig:fig7} report, respectively, the average $L_2$ distance between Models A and B and the cosine similarity of their denoising directions at each timestep.

Across training‐set sizes we observe a clear memorization-to-generalization trend.  
In small models ($\le1k$ images), at low noise ($\sigma\!\le\!0.01$) the
$L_2$ distance between Models A and B grows quickly and remains large, while
cosine similarity of their denoising directions collapses toward 0, consistent with
each model reverting to memorized training samples. When the corruption is large
($\sigma\!\ge\!0.2$) the same models still pull images back to the face manifold:
$L_2$ distances shrink and cosine similarity rises, showing that coarse structure
is shared even in the memorization regime.  
In large models ($\ge10k$ images), divergence at low noise is reduced
by an order of magnitude and the minimum cosine similarity plateaus around
$0.55\!-\!0.60$, indicating that the two independently trained models now share a
substantially aligned local score field.  At higher noise levels both metrics converge
to near-perfect agreement, matching the behavior reported in \cite{kadkhodaie2023}.

\subsection*{B3. Evaluation on high-dimensional synthetic datasets}

To assess whether our findings extend beyond images and 2D synthetic data, we additionally trained and investigated diffusion models on three \textit{high-dimensional} synthetic datasets:

\begin{itemize}
    \item \textsc{HD-Uniform}: A 100-dimensional Gaussian mixture with 5 isotropic components evenly spaced in Euclidean space.
    \item \textsc{HD-SharpCov}: A 100-dimensional Gaussian mixture with strongly anisotropic covariance matrices, testing sensitivity to sharp curvature.
    \item \textsc{ThinManifold}: A 2D manifold embedded in a 100D space with small variance in directions, designed to test how models behave on data following the thin manifold hypothesis.
\end{itemize}

\begin{figure}[h!]
    \centering
    \includegraphics[width=0.95\linewidth]{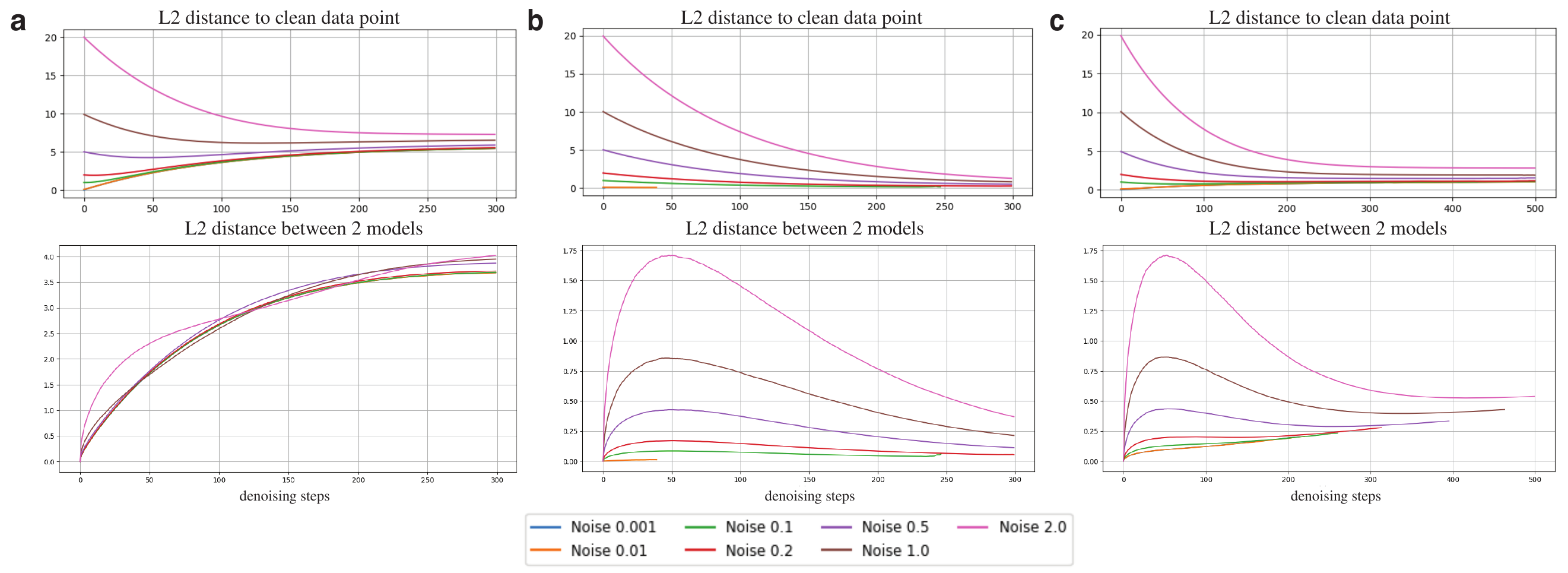}
    \caption{Trajectory divergence and denoising accuracy in high-dimensional datasets. Each column shows results for a different dataset: (a) \textsc{HD-Uniform}, (b) \textsc{HD-SharpCov}, and (c) \textsc{ThinManifold}. Top row: $L_2$ distance to the clean data point during denoising for a single model. Bottom row: $L_2$ distance between outputs of two models trained on disjoint subsets (Models A and B).}
    \label{fig:fig8}
\end{figure}

Fig. \ref{fig:fig8} compares denoising behavior across the three high-dimensional synthetic datasets. In the \textsc{HD-Uniform} mixture (left) the two models diverge the most: without geometric guidance they take different straight-line routes to the mode, so $L_2$ distances remain highest. In \textsc{HD-SharpCov} (middle) divergence is minimal—steep, elongated covariances funnel both models along the same principal axis. For the \textsc{ThinManifold} benchmark (right) divergence is slightly higher than SharpCov but still well below Uniform: the 2-D embedded surface guides trajectories, yet small curvature mismatches keep the models from perfect alignment.

\subsection*{B4. Replicating the Memorization-to-Generalization Transition in NCSN Models}

To verify that the results in Kadkhodaie et al. \cite{kadkhodaie2023} generalize to other diffusion models, we reproduced the transition from memorization to generalization using noise-conditional score networks (NCSN). Two NCSN models, A and B, are trained on disjoint subsets of CelebA at increasing dataset sizes. We then visualize the output of each model after denoising a common corrupted image with large noise initialization.

\begin{figure}[h!]
    \centering
    \includegraphics[width=0.95\linewidth]{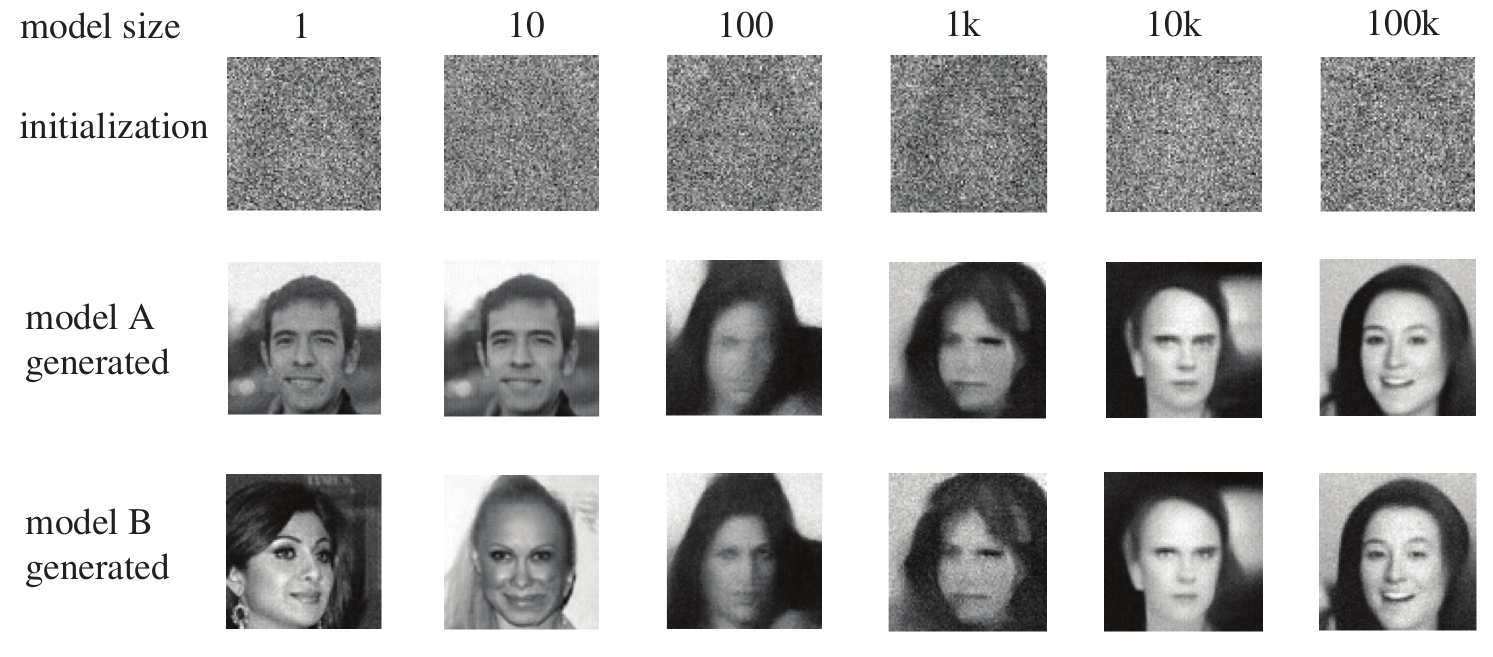}
    \caption{Outputs of Models A and B at high noise across training set sizes (1, 10, 100, 1k, 10k, 100k), showing transition from memorization to generalization. At small sizes, the two models produce divergent outputs resembling memorized training samples. As the training set grows, their outputs align first with weaker images and then with clear faces, indicating convergence to a shared estimate of the underlying score field.}
    \label{fig:fig9}
\end{figure}

\begin{figure}[h!]
    \centering
    \includegraphics[width=0.95\linewidth]{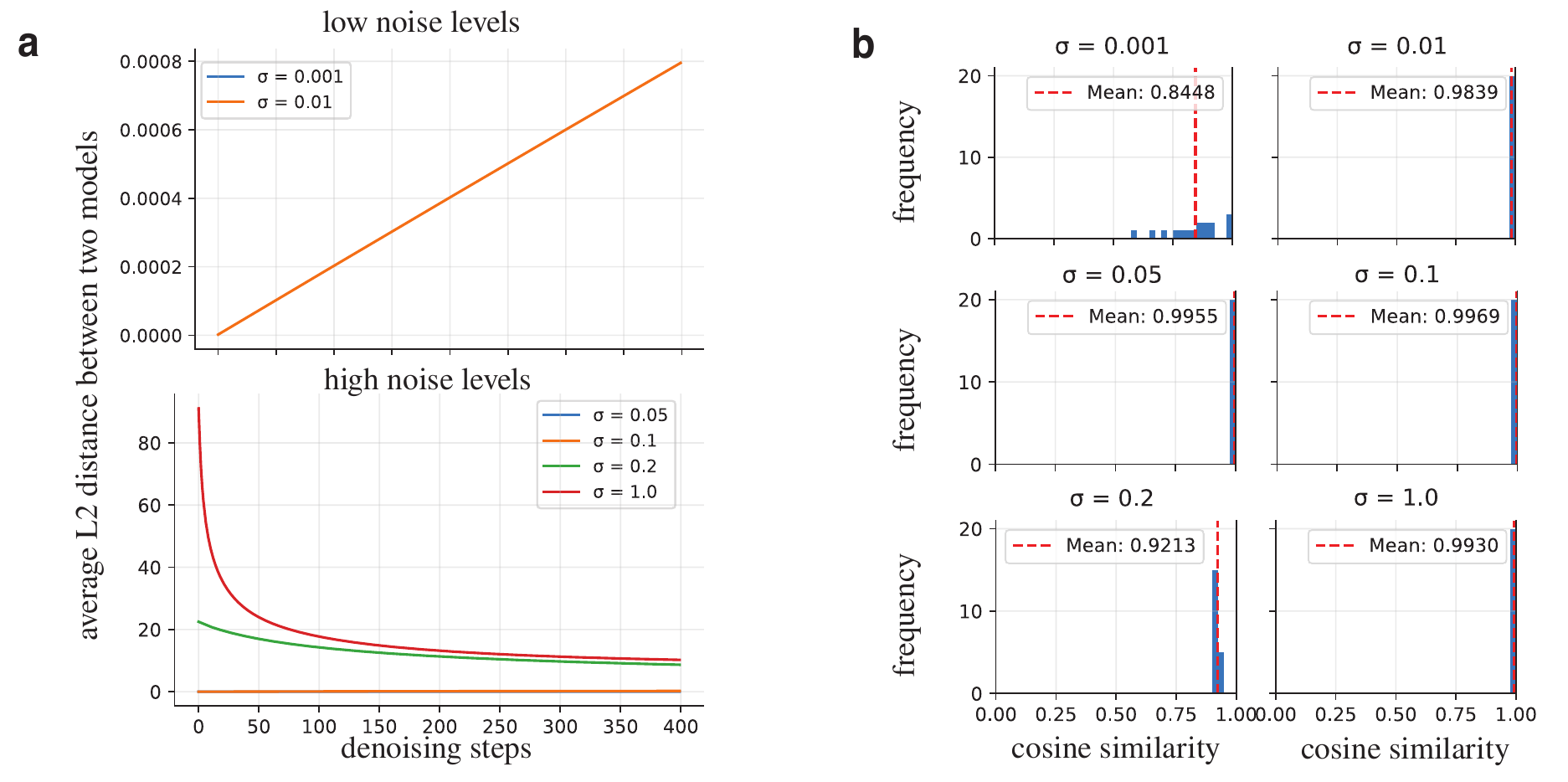}
    \caption{Probing the consistency of score-based diffusion models trained on disjoint subsets of CelebA under varying noise levels. (a) Average $L_2$ distance between denoised trajectories of Models A and B, separated into low and high noise ranges. (b) Histogram of cosine similarity of denoising direction vectors for Models A and B at different noise levels.}
    \label{fig:fig10}
\end{figure}

At the smallest training sizes (1, 10), the two models produce distinct outputs, suggesting reliance on memorized samples. By 10k–100k images, the denoised outputs of A and B converge, confirming generalization to a consistent score field. This result mirrors the original memorization-to-generalization transition and provides validation of our implementation.

When initializing the model at the vicinity of clean images (using the methodology from \ref{sec:exp:lownoise}) we see that NCSNs also experience divergence in low-noise settings. The $L_2$ distance between denoised trajectories of two models increases (see Fig. \ref{fig:fig10}a) and the cosine similarity of the predicted score vector decreases (see Fig. \ref{fig:fig10}b) with lower noise initializations, but less starkly than it does for the reconstruction UNet. This indicates that explicit score-matching may slightly improve local alignment but does not eliminate low-noise divergence.

\paragraph{Training details} 
\begin{itemize}
\item \textbf{Batch size}: 64 for CelebA images, 128 for synthetic GMMs
\item \textbf{Optimizer}: Adam, learning rate 0.001 for CelebA images and synthetic GMMs
\item \textbf{Training iterations}: data repeated until 200K datapoints reached per model and then trained for 100 epochs for CelebA; 500 epochs run for the synthetic datasets
\item \textbf{Input resolution}: $80 \times 80$ grayscale for CelebA images
\item \textbf{CelebA Subsets}: Subsets of size 1, 10, 100, 1k, 10k, 100k, sampled to ensure disjoint partitions for experiments involving model A and B.
\item \textbf{Synthetic datasets}: Gaussian Mixtures (\textsc{Uniform, SharpCov, Spiral, HD-Uniform, HD-SharpCov, ThinManifold}) each with up to 100k points.
\end{itemize}

\paragraph{Stopping criteria.} 
Reverse diffusion was run either until a fixed maximum number of steps (varies from 300 to 2000 based off model and dataset) or until convergence (when the change in predicted distance was below $10^{-2}$ between steps).

\section*{C. Computing resources}
\begin{itemize}
\item \textbf{Training hardware}: single NVIDIA A100 GPU (40GB)
\item \textbf{Training time}: 35 minutes per epoch per model for CelebA models; ~10 seconds per epoch per model for synthetic datasets
\item \textbf{Total compute}: 6 dataset sizes × 2 models x 100 epochs = 1200 total epochs for CelebA; 6 dataset sizes x 2 models x 6 synthetic distributions x 1000 epochs = 72000 epochs for synthetic datasets
\end{itemize}

\section*{D. Licenses for existing assets}
In our work, we leveraged several external assets that require proper attribution:
\begin{enumerate}
   \item \textbf{CelebA Dataset}: We used the CelebA dataset as described in Section 3.1. The dataset was introduced by Liu et al. \cite{liu2015} and is available at \url{http://mmlab.ie.cuhk.edu.hk/projects/CelebA.html} under a non-commercial research license. The dataset is available for non-commercial research purposes only.
   
   \item \textbf{UNet Denoiser Implementation}: This model follows the architecture described in Kadkhodaie et al. \cite{kadkhodaie2023} available at \url{https://github.com/LabForComputationalVision/memorization_generalization_in_diffusion_models} under the MIT License.
   
   \item \textbf{NCSN and SSM Implementation}: Both of these models follow the implementation described by Song et al. \cite{song2020} available at \url{https://github.com/ermongroup/ncsn} under the GPL-3.0 License.
\end{enumerate}

\section*{E. Code and data availability}
All code and experiment configurations used in this work are made publicly available at \url{https://github.com/lizardp1/diffusion_low_noise_regime}.

\end{document}